\documentclass[letterpaper]{article}
\usepackage{aaai18} 
\usepackage{graphicx}
\usepackage{subfigure}
\usepackage{multirow}
\usepackage{amsthm, amsmath, amssymb, amsxtra, amsfonts}
\usepackage{latexsym}
\usepackage{tikz}
\usepackage{wrapfig}
\usepackage [english] {babel}
\usepackage{lineno}
\usepackage{paralist}

\newtheorem{definition}{Definition}
\newtheorem{theorem}{Theorem}
\newtheorem{example}{Example}
\newtheorem{claim}{Claim}

\title{A Matrix Approach for Weighted Argumentation Frameworks: a Preliminary Report}
\author{
	Stefano Bistarelli \and Alessandra Tappini\\
	Universit\`a degli Studi di Perugia, Italy\\
	stafano.bistarelli@unipg.it\\
	alessandra.tappini@studenti.unipg.it\\
	\And
	Carlo Taticchi\\
	Gran Sasso Science Institute, Italy\\
	carlo.taticchi@gssi.it
}

\begin{document}

\maketitle

\begin{abstract}
	The assignment of weights to attacks in a classical Argumentation Framework allows to compute semantics by taking into account the different importance of each argument. We represent a Weighted Argumentation Framework by a non-binary matrix, and we characterize the basic extensions (such as $w$-admissible, $w$-stable, $w$-complete) by analysing sub-blocks of this matrix. Also, we show how to reduce the matrix into another one of smaller size, that is equivalent to the original one for the determination of extensions. Furthermore, we provide two algorithms that allow to build incrementally $w$-grounded and $w$-preferred extensions starting from a $w$-admissible extension.
\end{abstract}

\section{Introduction}\label{se:introduction}

An \emph{Abstract Argumentation Framework} (\emph{AF})~\cite{dung} is represented by a pair $\langle \mathcal{A}, R \rangle$ consisting of a set of arguments $\mathcal{A}$ and a binary relation of attack $R$ defined between some of them. 
Given a framework, it is possible to examine the question on which set(s) of arguments can be accepted, hence collectively surviving the conflict defined by $R$. Answering this question corresponds to define an argumentation semantics. 
The key idea behind \emph{extension-based} semantics is to identify some sets of arguments (called \emph{extensions}) that survive the conflict ``together''.
A very simple example of AF is  $\langle \{a,b\}, \{R(a,b), R(b,a)\} \rangle$, where two arguments $a$ and $b$ attack each other. In this case,  each of the two positions represented by either $\{a\}$ or $\{b\}$ can be intuitively valid, since no additional information is provided on which of the two attacks prevails. However, having weights on attacks results in such additional information, which can be fruitfully exploited in this direction. For instance, in case the attack $R(a,b)$ is stronger than (or preferred to) $R(b,a)$, taking the position defined by $a$ may result in a better choice for an intelligent agent, since it can be regarded as more reliable or relevant on the framework.

In a recent work, Xu and Cayrol represent an AF by a binary matrix and they give a characterization for stable, admissible and complete extensions by analysing sub-blocks of this matrix~\cite{xu15}. Also, they present the reduced matrix w.r.t.\ conflict-free subsets, by which the determination of extensions becomes more efficient, and that allows to determine $w$-grounded and $w$-preferred extensions.

Our aim is to extend the above mentioned results to Weighted Argumentation Frameworks (WAFs) by adopting the paradigm introduced in~\cite{DBLP:conf/cilc/BistarelliPS10,DBLP:conf/flairs/BistarelliRS16} for the semiring-based version of classical semantics. In particular, (i) we characterize $w$-conflict-free, $w$-admissible, $w$-stable and $w$-complete extensions by analysing sub-blocks of a non-binary matrix representing a given WAF, (ii) we show how to reduce this matrix to another one of smaller size that allows to more efficiently determine extensions, and (iii) we provide two algorithms that allow to build incrementally grounded and preferred extensions.

This paper is organized as follows: we first recall the basic definitions on AFs and on WAFs, then we give characterizations for weighted extensions by analysing the matrix associated with the given WAF. Finally, we present the matrix reductions of WAFs based on contraction and division of WAFs, and we provide methods for incrementally building $w$-grounded and $w$-preferred extensions.

\section{Weighted Argumentation Frameworks}\label{se:background}
In this section, we recollect the main definitions at the basis of AFs~\cite{dung}, and introduce c-semirings for dealing with attack-weights. We then rephrase some of the classical definitions, with the purpose to parametrise them with the notion of weighted attack and c-semiring. Last, we give definitions about the matrix representation for AFs.

\subsection{Abstract Argumentation Frameworks}\label{sec:argumentation}
\label{sec:argurelated} 
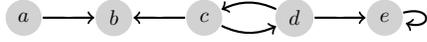
\begin{figure}[h]
	\centering
	\begin{tikzpicture}[scale=0.8, transform shape]
	\tikzstyle{every node} = [line width=1pt, shape=circle, fill=gray!35, minimum width=0.6cm]
	\node (a) at (3, 0) {$a$};
	\node (b) at +(0: 4.5) {$b$};
	\node (c) at +(0: 6) {$c$};
	\node (d) at +(0: 7.5) {$d$};
	\node (e) at +(0: 9) {$e$};
	\draw [line width = 0.8pt, ->] (a) -- (b) node[pos=.5, fill=white, above] {};
	\draw [line width = 0.8pt, ->] (c) -- (b) node[pos=.5, fill=white, above] {};
	\draw [line width = 0.8pt, ->] (d) -- (e) node[pos=.5, fill=white, above] {};
	\draw [line width = 0.8pt, ->] (c) edge[bend right=25]  (d);
	\draw [line width = 0.8pt, ->] (d) edge[bend right=25] (c);
	\draw [line width = 0.8pt, ->] (e) edge[loop right] (e);
	\end{tikzpicture}
	\caption{An example of AF.}\label{figure:argexample}
\end{figure}

In his pioneering work~\cite{dung}, Dung proposed \emph{Abstract Frameworks} for Argumentation, where (as shown in Figure~\ref{figure:argexample}) an argument is an abstract entity whose role is solely determined by its relations to other arguments:

\begin{definition}\label{def1}
	An Abstract Argumentation Framework (AF) is a pair $\langle \mathcal{A},R \rangle$ of a set $\mathcal{A}$ of arguments and a binary relation $R$ on $\mathcal{A}$, called attack relation. $\forall a_i, a_j \in \mathcal{A}$, $a_i R\, a_j$ (or $R(a_i, a_j)$) means that $a_i$ attacks $a_j$ ($R$ is asymmetric).
\end{definition}

Let $F = \langle\mathcal{A},R \rangle$ be an AF and $Z \subseteq A$. $R^+(Z)$ denotes the set of arguments attacked by $Z$ (a set $Z$ attacks a set $Z'$ if exist $a_i\in Z$ and $a_j\in Z'$ with $R(a_i, aj)$). $R^-(Z)$ denotes the set of arguments attacking $Z$. $I_{AF}$ denotes the set of arguments which are not attacked (also called initial arguments of $F$).

An \emph{argumentation semantics} is the formal definition of a method ruling the argument evaluation process. In the \emph{extension-based} approach, a semantics definition specifies how  to derive from an AF a set of extensions, where an extension $\mathcal{B}$ of an AF $\langle \mathcal{A},R \rangle$ is simply a subset of $\mathcal{A}$.
In Definition~\ref{def2e}  we define conflict-free sets:

\begin{definition}[Conflict-free]\label{def2e}
	A set $\mathcal{B} \subseteq \mathcal{A}$ is conflict-free
	iff no two arguments $a$ and $b$ in $\mathcal{B}$ exist
	such that $a$ attacks $b$. 
\end{definition}

All the following semantics rely (explicitly or implicitly) upon the concept of defence:

\begin{definition}[Defence~\cite{dung}]\label{def2}
	An argument $b$ is defended by a set $\mathcal{B} \subseteq
	\mathcal{A}$ (or $\mathcal{B}$ defends $b$) iff for any
	argument $a \in \mathcal{A}$, if $R(a,b)$ then
	$\exists c \in\mathcal{B}$ s.t., $R(c,a)$.
\end{definition}

\begin{definition}[Extension-based semantics]\label{def:semantics}
	\begin{compactitem}
	\item A conflict-free set $\mathcal{B} \subseteq \mathcal{A}$ is admissible iff each argument in $\mathcal{B}$ is defended by $\mathcal{B}$.
	\item An admissible extension $\mathcal{B} \subseteq \mathcal{A}$ is a complete extension iff each argument that is defended by $\mathcal{B}$ is in $\mathcal{B}$.
	\item A preferred extension is a maximal (w.r.t.\ set inclusion) admissible subset of $\mathcal{A}$.
	\item A grounded extension is a minimal (w.r.t.\ set inclusion) complete subset of $\mathcal{A}$.
	\item A conflict-free set $\mathcal{B} \subseteq \mathcal{A}$ is a stable extension iff for each argument which is not in $\mathcal{B}$, there exists an argument	in $\mathcal{B}$ that attacks it.
	\end{compactitem}
\end{definition}

\subsection{C-semirings}\label{sec:semirings}
C-semirings are \emph{commutative} ($\otimes$ is commutative) and \emph{idempotent} semirings (i.e., $\oplus$ is idempotent), where $\oplus$ defines a partial order $\leq_{\mathbb S}$. The obtained structure can be shown to be a complete lattice. 

\begin{definition}[c-semirings]
	A commutative semiring is a tuple $\mathbb S= \langle S,\oplus,\otimes,$ $\bot,\top \rangle$  such that $S$ is a set, $\top, \bot \in S$, and $\oplus, \otimes : S \times S \rightarrow S$
	are binary operators making the triples $\langle S, \oplus, \bot \rangle$ and $\langle S, \otimes, \top \rangle$
	commutative monoids (semi-groups with identity), satisfying \emph{i)}  $\forall s, t, u \in S.s \otimes (t \oplus u) = (s \otimes t) \oplus (s \otimes u)$ (distributivity), and \emph{ii)}  $\forall s \in S.s \otimes \bot = \bot$ (annihilator). If $\forall s,t \in S. s\oplus(s\otimes t) = s$, the semiring is said to be absorptive.
\end{definition}

Well-known instances of c-semirings are:
\begin{compactitem}
	\item $\mathbb S_\mathit{boolean}= \langle \{\mathit{false}, \mathit{true}\}, \vee, \wedge, \mathit{false},\mathit{true}\rangle$\footnote{Boolean c-semirings can be used to model crisp problems and classical Argumentation~\cite{dung}.}, 
	\item $\mathbb S_\mathit{fuzzy}=$ $\langle [0,1],\max, \min, 0,$  $ 1 \rangle$, 
	\item $\mathbb S_\mathit{bottleneck}  = \langle \mathbb{R}^+ \cup\{+\infty\},  \max, \min,$ $0, \infty \rangle$,
	\item $\mathbb S_\mathit{probabilistic}= \langle [0,1],$ $\max, \times, 0, 1 \rangle$, 
	\item $\mathbb S_\mathit{weighted}= \langle \mathbb{R}^+ \cup\{+\infty\}, min, +, +\infty, 0 \rangle$.
\end{compactitem}

C-semirings provide a structure that reveals to be suitable for Weighted Argumentation Frameworks. In fact, values in $S$ can be used as weights for relations, while the operators $\oplus$ and $\otimes$ allow to define an ordering among weights.

\subsection{Weighted AFs}\label{sec:wafs}
The following definition reshapes the notion of Weighted Argumentation Framework into \emph{semiring-based WAF}, called $\mathit{WAF}_{\mathbb S}$:

\begin{definition}[Semiring-based WAF]\label{def:waf}
	A semiring-based WAF ($\mathit{WAF}_{\mathbb S}$) is a quadruple
	$\langle \mathcal{A}, R, W, \mathbb S \rangle$, where $\mathbb S$ is a c-semiring $
	\langle S, \oplus, \otimes, \bot, \top \rangle$, $\mathcal{A}$ is a set
	of arguments, $R$ the attack binary-relation  on
	$\mathcal{A}$, and $W: \mathcal{A} \times \mathcal{A}
	\longrightarrow S$ is a binary function. Given $a,b \in \mathcal{A}$ and $R(a, b)$, then $W(a, b) = s$ means that $a$ attacks $b$ with a weight
	$s \in S$. Moreover, we require that $R(a,b)$ iff $W(a,b) <_{\mathbb S} \top$.
\end{definition}

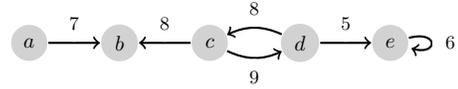
\begin{figure}[h]
	\centering
	\begin{tikzpicture}[scale=0.8, transform shape]
	\tikzstyle{every node} = [line width=1pt, shape=circle, fill=gray!35, minimum width=0.6cm]
	\node (a) at (3, 0) {$a$};
	\node (b) at +(0: 4.5) {$b$};
	\node (c) at +(0: 6) {$c$};
	\node (d) at +(0: 7.5) {$d$};
	\node (e) at +(0: 9) {$e$};
	\draw [line width = 0.8pt, ->] (a) -- (b) node[pos=.5, fill=white, above] {\footnotesize{7}};
	\draw [line width = 0.8pt, ->] (c) -- (b) node[pos=.5, fill=white, above] {\footnotesize{8}};
	\draw [line width = 0.8pt, ->] (d) -- (e) node[pos=.5, fill=white, above] {\footnotesize{5}};
	\draw [line width = 0.8pt, ->] (c)  edge[bend right=25]   node[fill=white, below] {\footnotesize{9}} (d);
	\draw [line width = 0.8pt, ->] (d) edge[bend right=25] node[fill=white, above] {\footnotesize{8}} (c);
	\draw [line width = 0.8pt, ->] (e) edge[loop right] node[fill=white, right] {\footnotesize{6}} (e);
	\end{tikzpicture}
	\vspace{-0.3cm}
	\caption{An example of  WAF, adding weights to Figure~\ref{figure:argexample}.}\label{fig:argnetex}
\end{figure}

In Figure~\ref{fig:argnetex}, we provide an example of a WAF describing the $\mathit{WAF}_{\mathbb S}$ defined by
$\mathcal{A}= \{a, b, c, d, e\}$, $R= \{(a, b), (c, b), (c, d), (d, c), (d, e),$ $(e, e)\}$,  with $W(a, b)= 7, W(c, b)= 8, W(c, d)= 9, W(d, c)= 8, W(d, e)= 5, W(e, e)= 6$, and $\mathbb S= \langle \mathbb{R}^+ \cup \{\infty\}, \min, +, \infty, 0\rangle$ (i.e., the weighted semiring). 


Therefore, each attack  is associated with a semiring value that represents the ``strength'' of an attack between two arguments.
We can consider the weights in Figure~\ref{fig:argnetex} as supports to the associated attack, as similarly suggested in \cite{DBLP:journals/ai/DunneHMPW11}. A semiring value equal to the top element of the c-semiring $\top$ (\emph{e.g.}, $0$ for the weighted semiring) represents a no-attack relation between two arguments. On the other side, the bottom element, i.e., $\bot$ (\emph{e.g.}, $\infty$ for the weighted semiring), represents the strongest attack possible. In the following, we will use $\bigotimes$ to indicate the $\otimes$ operator of the c-semiring $\mathbb S$ on a set of  values:

\begin{definition}[Attacks to/from sets of arguments]\label{def:wset}
	Let  $\mathit{WF}= \langle \mathcal{A}, R, W, \mathbb S \rangle$ be a $\mathit{WAF}_{\mathbb S}$. A set of arguments $\mathcal{B}$ attacks a set of arguments $\mathcal{D}$ and the weight of such attack is  $k \in S$, if $$W(\mathcal{B},\mathcal{D})= \displaystyle\bigotimes_{b \in \mathcal{B}, d \in \mathcal{D}} W(b,d) = k.$$
\end{definition}

For example, looking at Figure~\ref{fig:argnetex}, we have that $W(\{a, c\}, b) = 15$, $W(c,\{b,d\}) = 17$, and $W(\{a,c\}, \{b,d\}) = 24$.




\begin{definition}[$w$-defence~\cite{DBLP:conf/flairs/BistarelliRS16}]\label{defw2}
	Given a $\mathit{WAF}_{\mathbb S}$,  $\mathit{WF}= \langle \mathcal{A}, R, W, \mathbb S \rangle$,
	$\mathcal{B} \subseteq
	\mathcal{A}$ $w$-defends $b \in \mathcal{A}$  iff $\forall a \in \mathcal{A}$ such that $R(a,b)$,  we have that $W(a, \mathcal{B} \cup \{b\}) \geq_{\mathbb S} W(\mathcal{B}, a)$.
\end{definition}

A set $\mathcal{B} \subseteq \mathcal{A}$ $w$-defends an argument $b$ from $a$, if the $\otimes$  of all attack weights from $\mathcal{B}$ to $a$ is worse\footnote{When considering the partial order of a generic semiring, we use ``worse'' or ``better'' because ``greater'' or ``lesser'' would be misleading: in the weighted semiring, $7 \leq_{\mathbb S} 3$, i.e., lesser means better.} (w.r.t.\ $\leq_{\mathbb S}$) than the $\otimes$ of the attacks from $a$ to $\mathcal{B} \cup \{b\}$. For example, the set $\{c\}$ in Figure~\ref{fig:argnetex} defends $c$ from $d$ because $W(d, \{c\}) \geq_{\mathbb S} W(\{c\}, d)$, i.e., ($8 \leq 9$).

\begin{definition}[$w$-conflict-free]\label{cfw}Given a $\mathit{WAF}_{\mathbb S}$ $\mathit{WF} = \langle \mathcal{A}, R, W,\mathbb S \rangle$, a subset of arguments $\mathcal{B} \subseteq \mathcal{A}$ is $w$-conflict-free if $W(\mathcal{B}, \mathcal{B}) = \top$.
\end{definition}

\begin{definition}[$w$-admissible]\label{admw}Given a $\mathit{WAF}_{\mathbb S}$ $\mathit{WF}= \langle \mathcal{A}, R, W, \mathbb S \rangle$, a $w$-conflict-free set $\mathcal{B} \subseteq \mathcal{A}$ is $w$-admissible iff the arguments in $\mathcal{B}$ are $w$-defended by $\mathcal{B}$ from the arguments in $\mathcal{A} \setminus \mathcal{B}$.
\end{definition}

\begin{definition}[$w$-complete]\label{comw}A $w$-admissible extension $\mathcal{B} \subseteq \mathcal{A}$ is also a $w$-complete extension iff each argument $b \in \mathcal{A}$ such that $\mathcal{B} \cup \{b\}$ is $w$-admissible belongs to $\mathcal{B}$, i.e., $b \in \mathcal{B}$.
\end{definition}

\begin{definition}[$w$-preferred and $w$-grounded]A $w$-preferred extension is a maximal (w.r.t.\ set inclusion) $w$-admissible subset of $\mathcal{A}$. The least (w.r.t.\ set inclusion) $w$-complete extension is the $w$-grounded extension.
\end{definition}

\begin{definition}[$w$-stable]\label{staw}Given $\mathit{WF}= \langle \mathcal{A}, R, W, \mathbb S \rangle$, a $w$-admissible set $\mathcal{B}$ is also a $w$-stable extension iff  $ \forall a \notin \mathcal{B}, \exists b \in \mathcal{B}$ such that $W(b,a) \leq_{\mathbb S} \top$.
\end{definition}

\section{The Matrix Representation for  WAFs}\label{sse:matrix-weighted}
Given an AF $F$, we can obtain a matrix representing $F$ by using Definition 4 in~\cite{xu15}. We extend this definition to represent WAFs through matrices.

\begin{definition}\label{matrix-weighted}
	Let $F=\langle \mathcal{A}, R, W, \mathbb S \rangle$ be a WAF with $\mathcal{A} = \{1,2,\dots,n\}$. The matrix of $F$ corresponding to the permutation $(i_1, i_2,\dots ,i_n)$ of $\mathcal{A}$, denoted by $M(i_1, i_2,\dots,i_n)$, is a matrix of order $n$, its elements being determined by the following rules: $(1)$ $a_{s,t} = w$ iff $(i_s, i_t) \in R$ and $W(i_s,i_t)=w$;  $(2)$ $a_{s,t} = \top$ iff $(i_s, i_t) \notin R$.
\end{definition}

\begin{figure}[h]
	\centering
	\begin{tikzpicture}[scale=0.8, transform shape]
	\tikzstyle{every node} = [line width=1pt, shape=circle, fill=gray!35, minimum width=0.7cm]
	\node (a) at (0, 0) {$a$};
	\node (b) at (2, 0) {$b$};
	\node (c) at (4, 0) {$c$};
	\draw [line width = 0.8pt, ->] (a) edge[bend right=25] node[pos=.5, fill=white, below] {\footnotesize{7}} (b);
	\draw [line width = 0.8pt, ->] (b) edge[bend right=25] node[pos=.5, fill=white, above] {\footnotesize{9}} (a);
	\draw [line width = 0.8pt, ->] (c) -- (b) node[pos=.5, fill=white, below] {\footnotesize{8}};
	
	\end{tikzpicture}
	\vspace{-0.3cm}
	\caption{\label{fi:ex2}Example of a WAF with $\mathbb S = \mathbb S_\mathit{weighted}$.}
\end{figure}
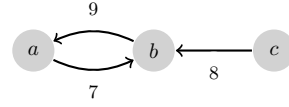

\begin{example}\label{ex2}
	Given $F=\langle \mathcal{A}, R, W, \mathbb S \rangle$ as in Figure~\ref{fi:ex2}. The matrices of $F$ corresponding to the permutations $(a,b,c)$ and $(a,c,b)$ are
	\[
	\bordermatrix{
		~ & a & b & c \cr
		a & 0 & 7 & 0 \cr
		b & 9 & 0 & 0 \cr
		c & 0 & 8 & 0
	}
	~~and~~
	\bordermatrix{
		~ & a & c & b \cr
		a & 0 & 0 & 7 \cr
		c & 0 & 0 & 8 \cr
		b & 9 & 0 & 0
	}
	\]
\end{example}

\section{Characterizing extensions of a WAF}\label{se:characterizations}
In this section, we mainly focus on the characterization of various extensions in the matrix $M(AF)$ representing a WAF.

	\subsection{Characterizing the w-conflict-free subsets}\label{sse:conflict-free}
	
	The basic requirement for extensions is conflict-freeness. So, we will discuss the matrix condition which insures that a subset of a WAF is conflict-free.
	
	\begin{definition}\label{cf-matrix}
	Let $F=\langle \mathcal{A}, R, W, \mathbb S \rangle$ be a WAF with $\mathcal{A} = \{1,2,\dots,n\}$ and $Z=(i_1, i_2, \dots, i_k) \subseteq \mathcal{A}$. The $k \times k$ sub-block \\
	\[M_{i,j}=
	\begin{pmatrix}
		a_{i_1,i_1} & a_{i_1,i_2} & \dots & a_{i_1,i_k} \\
		a_{i_2,i_1} & a_{i_2,i_2} & \dots & a_{i_2,i_k} \\
		\vdots & \vdots & \ddots & \vdots \\
		a_{i_k,i_1} & a_{i_k,i_2} & \dots & a_{i_k,i_k}
	\end{pmatrix}\]\\
	of $M(AF)$ is called the cf-sub-block of $Z$, and denoted by $M^{cf}(Z)$ for short. We use this sub-block to find conflict-free subsets of arguments.
	\end{definition}

	\begin{claim}\label{w-cf}
		Given $F=\langle \mathcal{A}, R, W, \mathbb S \rangle$ with $\mathcal{A} = \{1,2,\dots,n\}$, $Z=(i_1, i_2, \dots, i_k) \subseteq \mathcal{A}$ is $w$-conflict-free iff all the elements in the cf-sub-block $M^{cf}(Z)$ are $\top$.
	\end{claim}

	\begin{example} 
		Consider the WAF of Figure~\ref{fi:ex2}. We have that
		$M^{cf}(\{a,c\}) =
			\begin{pmatrix}
				0 & 0 \\
				0 & 0 \\
			\end{pmatrix}$,
		$M^{cf}(\{a,b\}) =
			\begin{pmatrix}
			0 & 7  \\
			9 & 0 \\
			\end{pmatrix}$
		and $M^{cf}(\{b,c\}) =
			\begin{pmatrix}
			0 & 0 \\
			8 & 0
			\end{pmatrix}$.
		By Theorem~\ref{w-cf}, $\{a, c\}$ is $w$-conflict-free, while $\{a, b\}$ and $\{b, c\}$ are not.
	\end{example}

	\subsection{Characterizing the w-admissible subsets}\label{sse:admissible}
	
	From Definition~\ref{admw}, we know that arguments belonging to a $w$-admissible subset $\mathcal{B} \subseteq \mathcal{A}$ are $w$-defended from the arguments in $\mathcal{A} \setminus \mathcal{B}$.
	
	\begin{definition}\label{s-matrix}
		Let $F=\langle \mathcal{A}, R, W, \mathbb S \rangle$ be a WAF with $\mathcal{A} = \{1,2,\dots,n\}$, $Z=(i_1, i_2, \dots, i_k) \subseteq \mathcal{A}$ and $\mathcal{A}\setminus Z=\{j_1,j_2,\dots, j_h\}$. The $k \times h$ sub-block \\
		\[M^{i_1, i_2, \dots, i_k}_{j_1,j_2,\dots, j_h}=
		\begin{pmatrix}
		a_{i_1,j_1} & a_{i_1,j_2} & \dots & a_{i_1,j_h} \\
		a_{i_2,j_1} & a_{i_2,j_2} & \dots & a_{i_2,j_h} \\
		\vdots & \vdots & \ddots & \vdots \\
		a_{i_k,j_1} & a_{i_k,j_2} & \dots & a_{i_k,j_h}
		\end{pmatrix}\]\\
		of $M(AF)$ is called the s-sub-block of $Z$, and denoted by $M^{s}(Z)$ for short. The $h \times k$ sub-block of $M(AF)$\\
		\[M^{j_1,j_2,\dots, j_h}_{i_1, i_2, \dots, i_k}=
		\begin{pmatrix}
		a_{j_1,i_1} & a_{j_1,i_2} & \dots & a_{j_1,i_k} \\
		a_{j_2,i_1} & a_{j_2,i_2} & \dots & a_{j_2,i_k} \\
		\vdots & \vdots & \ddots & \vdots \\
		a_{j_h,i_1} & a_{j_h,i_2} & \dots & a_{j_h,i_k}
		\end{pmatrix}\]\\
		is called the $\overline{s}$-sub-block\footnote{In~\cite{xu15}, $M^{\overline{s}}$ is denoted as $M^{a}$ and it is called the a-sub-block.} of $Z$, and denoted by $M^{\overline{s}}(Z)$.
	\end{definition}
	
	\begin{theorem}\label{w-adm}
		Given $F=\langle \mathcal{A}, R, W, \mathbb S \rangle$ with $\mathcal{A} = \{1,2,\dots,n\}$, a $w$-conflict-free subset $Z=\{i_1, i_2, \dots, i_k\} \subseteq \mathcal{A}$ is $w$-admissible iff $\forall j_q \in \mathcal{A} \setminus Z$, $\displaystyle\bigotimes_{i \in Z} W(i,j_q) \leq_{\mathbb S} \displaystyle\bigotimes_{i \in Z} W(j_q,i)$, where $W(i,j_q)$ refers to the column vector $M^s_{*,q}$ of the s-sub-block $M^s(Z)$ and $W(j_q,i)$ refers to the column vector $M^{\overline{s}}_{*,q}$ of the $\overline{s}$-sub-block $M^{\overline{s}}(Z)$.
	\end{theorem}
	
	\begin{example} 
		Let's consider the $w$-conflict-free subsets $\{a\}$ and $\{a,c\}$ (see Figure~\ref{fi:ex2}). We have
		$M^{s}(\{a\}) =
		\begin{pmatrix}
		7 & 0 \\
		\end{pmatrix}$ and
		$M^{\overline{s}}(\{a\}) =
		\begin{pmatrix}
		9\\
		0
		\end{pmatrix}$, the weight associated to the column vector $M^s_{*,1}$ of $M^s(\{a\})$ is $W(a,b) = 7$ while the weight associated to the row vector $M^{\overline{s}}_{1,*}$ of $M^{\overline{s}}(\{a\})$ is $W(b,a) = 9$. Since $7 \geq_{\mathbb S} 9$, $\{a\}$ is not $w$-admissible in $F$ according to Theorem~\ref{w-adm}.\\However, from
		$M^{s}(\{a,c\}) =
		\begin{pmatrix}
		7\\
		8
		\end{pmatrix}$ and
		$M^{\overline{s}}(\{a,c\}) =
		\begin{pmatrix}
		9 & 0
		\end{pmatrix}$, we know that the weight associated to the column vector $M^s_{*,1}$ of $M^s(\{a,c\})$ is $W(a,b) \otimes W(c,b) = 7+8 = 15$ while the weight associated to the row vector $M^{\overline{s}}_{1,*}$ of $M^{\overline{s}}(\{a,c\})$ is $W(b,a) \otimes  W(b,c) = 9+0 = 9$. Since $15 \leq_{\mathbb S} 9$, we claim that $\{a,c\}$ is $w$-admissible in $F$ by Theorem~\ref{w-adm}.
	\end{example}
	
\subsection{Characterizing the w-stable extensions}\label{sse:stable}

We can say whether a $w$-admissible subset $\mathcal{B} \subseteq \mathcal{A}$ is also a $w$-stable extension by checking if all arguments in $\mathcal{A} \setminus \mathcal{B}$ are attacked by arguments in $\mathcal{B}$. On this purpose, we can use the already defined matrix $M^{s}(Z)$.
	
	\begin{theorem}\label{w-stb}
		Given $F=\langle \mathcal{A}, R, W, \mathbb S \rangle$ with $\mathcal{A} = \{1,2,\dots,n\}$, a $w$-admissible subset $Z=\{i_1, i_2, \dots, i_k\} \subseteq \mathcal{A}$ is a $w$-stable extension iff each column vector of the s-sub-block $M^{s}(Z)$ of $M(AF)$ contains only elements different from $\top$, where $\{j_1,j_2,\dots, j_h\}$ is a permutation of $\mathcal{A}\setminus Z$.
	\end{theorem}
	
	\begin{example} 
		Let's consider the $w$-admissible subset $\{a, c\}$ (see Figure~\ref{fi:ex2}). Since the only column vector of
		$M^{s}(\{a,c\}) =
		\begin{pmatrix}
		7\\
		8\\
		\end{pmatrix}$ contains some elements different from $\top$, we claim that $\{a, c\}$ is a $w$-stable extension of $F$, according to Theorem~\ref{w-stb}.
	\end{example}

\subsection{Characterizing the w-complete extensions}\label{sse:complete}

From the definition of $w$-complete extension, it comes that in addition of considering relations between arguments all inside $\mathcal{B}$ and between arguments in $\mathcal{B}$ and those outside $\mathcal{B}$, we also need to take into account attacks thoroughly outside $\mathcal{B}$. We give the following definition and theorem.

	\begin{definition}\label{c-matrix}
		Let $F=\langle \mathcal{A}, R, W, \mathbb S \rangle$ be a WAF with $\mathcal{A} = \{1,2,\dots,n\}$, $Z=(i_1, i_2, \dots, i_k) \subseteq \mathcal{A}$ and $\mathcal{A}\setminus Z=\{j_1,j_2,\dots, j_h\}$.The $h \times h$ sub-block\\
		\[M^{j_1, j_2, \dots, j_h}_{j_1,j_2,\dots, j_h}=
		\begin{pmatrix}
		a_{j_1,j_1} & a_{j_1,j_2} & \dots & a_{j_1,j_h} \\
		a_{j_2,j_1} & a_{j_2,j_2} & \dots & a_{j_2,j_h} \\
		\vdots & \vdots & \ddots & \vdots \\
		a_{j_h,j_1} & a_{j_h,j_2} & \dots & a_{j_h,j_h}
		\end{pmatrix}\]\\
		of $M(AF)$ is called the c-sub-block of $Z$, and denoted by $M^{c}(Z)$ for short.
	\end{definition}

	\begin{theorem}\label{w-cmp}
		Given $F=\langle \mathcal{A}, R, W, \mathbb S \rangle$ with $\mathcal{A} = \{1,2,\dots,n\}$, a $w$-admissible subset $Z=\{i_1, i_2, \dots, i_k\} \subseteq \mathcal{A}$ is $w$-complete iff
	\begin{compactenum}
		\item[(1)] if some column vector $M^s_{*,p}$ of the s-sub-block $M^s(Z)$ contains only $\top$ elements, then its corresponding column vector $M^c_{*,p}$ of the c-sub-block $M^c(Z)$ contains some element different from $\top$ and
		\item[(2)] for each column vector $M^c_{*,p}$ of the c-sub-block $M^c(Z)$ appearing
		in (1), which contains some element different from $\top$, there is at least one element $a_{j_q,j_p} \neq \top$ of $M^c_{*,p}$ such that $\displaystyle\bigotimes_{i \in Z} W(j_q,i) \otimes W(j_q,j_p) \leq_{\mathbb S} \displaystyle\bigotimes_{i \in Z} W(i,j_q) \otimes W(j_p,j_q)$, where $W(i,j_q)$ refers to the column vector $M^s_{*,q}$ of the s-sub-block $M^s(Z)$, where $W(j_q,i)$ refers to the column vector $M^{\overline{s}}_{*,q}$ of the s-sub-block $M^{\overline{s}}(Z)$, $\{j_1,j_2,\dots, j_h\} = \mathcal{A}\setminus Z$ and $1 \leq q,p \leq h$.
	\end{compactenum}
	\end{theorem}

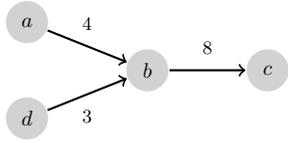
\begin{figure}[h]
		\centering
		\begin{tikzpicture}[scale=0.8, transform shape]
		\tikzstyle{every node} = [line width=1pt, shape=circle, fill=gray!35, minimum width=0.7cm]
		\node (a) at (0, 0) {$a$};
		\node (b) at (2, -0.8) {$b$};
		\node (c) at (4, -0.8) {$c$};
		\node (d) at (0, -1.6) {$d$};
		\draw [line width = 0.8pt, ->] (a) -- (b) node[pos=.5, fill=none, above] {\footnotesize{4}};
		\draw [line width = 0.8pt, ->] (b) -- (c) node[pos=.5, fill=none, above] {\footnotesize{8}};
		\draw [line width = 0.8pt, ->] (d) -- (b) node[pos=.5, fill=none, below] {\footnotesize{3}};
		\end{tikzpicture}
		\vspace{0.2cm}
		\caption{\label{fi:ex3}Example of a WAF with $\mathbb S = \mathbb S_\mathit{weighted}$.}
	\end{figure}

	\begin{example}\label{ex3} 
		Given $F=\langle \mathcal{A}, R, W, \mathbb S \rangle$ as in Figure~\ref{fi:ex3}. According to Definition~\ref{matrix-weighted}, the matrix of $F$ is as follows
		\[M(AF) =
		\begin{pmatrix}
		0 & 4 & 0 & 0 \\
		0 & 0 & 8 & 0 \\
		0 & 0 & 0 & 0 \\
		0 & 3 & 0 & 0
		\end{pmatrix}\]
		By Theorem~\ref{w-adm}, we have that $Z=\{a,d\}$ is $w$-admissible. Note that the matrix
		$M^s(\{a,d\})=
		\begin{pmatrix}
			4 & 0\\
			3 & 0
		\end{pmatrix}$ has a column vector 
		$M^s_{*,2} =
		\begin{pmatrix}
		0\\
		0
		\end{pmatrix}$ corresponding in 
		$M^c(\{a,d\})=
		\begin{pmatrix}
		0 & 8\\
		0 & 0
		\end{pmatrix}$ to the column vector 
		$M^c_{*,2} =
		\begin{pmatrix}
		8\\
		0
		\end{pmatrix}$. For $a_{b,c}=8$ in $M^c_{*,2}$, the corresponding column vector $M^s_{*,1}$ in $M^s(\{a,d\})$ has $W(a,b) \otimes W(d,b) = 4+3 = 7$. Since $8 \leq_{\mathbb S} 7$, according to Theorem~\ref{w-cmp}, we claim that $\{a,d\}$ is a $w$-complete extension of $F$.
	\end{example}

\section{Matrix reduction for WAFs}\label{se:reduction}

	Most of the time, it is convenient to reduce the size of the matrix before performing further operations on it. Below, we provide a method to contract the $w$-conflict-free subset of a matrix into a single entity, without affecting the computation of the extensions. Moreover, we show an iterative procedure for building $w$-grounded and $w$-preferred extensions.

	\subsection{Matrix reduction by contraction}\label{sse:contraction}
	Starting from a conflict-free sub-block, we can characterize $w$-admissible, $w$-stable and $w$-complete extensions of a WAF. Contracting such a sub-block, we obtain a new matrix of smaller size, but with the same semantics status as the original one.
	
	\begin{definition}
		Let $M(AF)$ be the matrix of a WAF. The combination of two rows $i$ and $j$ of the matrix $M(AF)$ consists in ``combining'' the elements in the same position of the rows. If $w_i$ and $w_j$ are elements in the same position of the rows $i$ and $j$ respectively, their combination is given by the rule $w_i \otimes w_j$. The combination of two columns of the matrix $M(AF)$ is similar as the combination of two rows.
	\end{definition}

	For a $w$-conflict-free subset $Z = {i_1, i_2, \dots, i_k}$, we can contract the sub-block $M^{cf}(Z)$ into a single entry in the matrix. This new entry will have the same status as $M^{cf}(Z)$ w.r.t.\ the extension-based semantics. Thus the matrix $M(AF)$ can be reduced into another matrix $M^r_Z(AF)$ with order $n-k+1$ by applying the following rules: let $1 \leq t \leq k$, for each $s$ such that $1 \leq s \leq k$ and $s \neq t$,
	
	\begin{compactenum}
		\item combine rows $i_s$ to the row $i_t$;
		\item combine column  $i_s$ to the column $i_t$;
		\item delete row  $i_s$ and column $i_s$.
	\end{compactenum}

	The matrix $M^r_Z(AF)$ obtained in this way is called the reduced matrix w.r.t.\ the conflict-free subset $Z$. Also, the original WAF can be reduced into a new one with $n-k+1$ arguments by applying the following rules. Let $A \setminus Z = \{j_1, j_2, \dots, j_h\}$ and $1 \leq t \leq k$. For each $s$ such that $1 \leq s \leq k$ and $s \neq t$, and each $q$ such that $1 \leq q \leq h$, set $W((i_t,j_q)) = 0$ and $W((j_q,i_t)) = 0$. Then,

	\begin{compactenum}
		\item if $(i_s,j_q) \in R$, combine $(i_t,j_q)$ to $R$ and set $W((i_t,j_q)) = W((i_t,j_q)) \otimes W((i_s,j_q))$;
		\item if $(j_q,i_s) \in R$, combine $(j_q,i_t)$ to $R$ and set $W((j_q,i_t)) = W((j_q,i_t)) \otimes W((j_q,i_s))$;
		\item delete $(i_s,j_q)$ and $(j_q,i_s)$ from $R$.
	\end{compactenum}

	Let $R^r_Z$ denote the new relation and $A^r_Z = \{i_t\} \cup (A \setminus Z)$, then $(A^r_Z,R^r_Z)$ is a new AF called the \emph{reduced} AF w.r.t.\ $Z$. Obviously, the reduced matrix $M^r_Z(AF)$ is exactly the matrix obtained from $A^r_Z$ and $R^r_Z$.

	\begin{theorem}
		Given $F=\langle \mathcal{A}, R, W, \mathbb S \rangle$ with $A = {1,2,\dots,n}$, let $Z = \{i_1,i_2,\dots,i_k\} \subseteq A$ be conflict-free and $1 \leq t \leq k$. Then $Z$ is stable (resp. admissible, complete, preferred) in AF iff $\{i_t\}$ is stable (respectively admissible, complete, preferred) in the reduced $F=\langle \mathcal{A}^r_Z, R^r_Z, W, \mathbb S \rangle$.
	\end{theorem}
	
	\subsection{Matrix reduction by division}\label{sse:division}
	Let $F = \langle \mathcal{A}, R, W, \mathbb S \rangle$ be a WAF. The $w$-grounded extension of $F$ can be viewed as the union of two subsets $I_{AF}$ and $E$: $I_{AF}$ consists of the initial arguments of $F$ and $E$ is the $w$-grounded extension, $w$-defended by $F$, of the remaining sub-AF w.r.t.\ $I_{AF}$ (that is $F\mid_B$, where $B = \mathcal{A} \setminus (I_{AF} \cup R^+(I_{AF}))$). On the other hand, a $w$-preferred extension coincides with an admissible extension $E$, $w$-defended by $F$, from which the associated remaining sub-AF $F \mid_C$ (where $C = \mathcal{A} \setminus (E \cup R^+(E))$) has no nonempty admissible extension. We have the following theorem.
	
	\begin{theorem}\label{th:union}
		Let $F=\langle \mathcal{A}, R, W, \mathbb S \rangle$ be a WAF, $Z \subseteq A$ be a $w$-admissible extension of $F$, and $B = \mathcal{A} \setminus (Z \cup R^+(Z))$. If $T \subseteq B$ is a $w$-admissible (resp. $w$-stable, $w$-complete, $w$-preferred) extension, $w$-defended by $F$, of the remaining sub-AF w.r.t.\ $Z$ ($F \mid _B$), then $Z \cup T$ is a $w$-admissible (resp. $w$-stable, $w$-complete, $w$-preferred) extension of $F$.
	\end{theorem}

	\begin{example}
		Given $F$ in Figure~\ref{fi:ex3}, consider $Z = \{a\}$ and $T = \{d\}$, with $T \subset B = \mathcal{A} \setminus (Z \cup R^+(Z)) = \{c,d\}$. $Z$ is $w$-admissible in $F$ and $T$ is $w$-admissible in $F \mid _B$. Then, for Theorem~\ref{th:union}, $Z \cup T = \{a,d\}$ is a $w$-admissible extension of $F$.
	\end{example}
	
	\subsection*{Building w-grounded extensions}
	A $w$-grounded extension can be built incrementally by starting from a $w$-admissible extension. Let $I_1$ be the set of initial arguments of $F$, then $I_1$ is a $w$-admissible extension. If $F$ has no initial arguments, then the $w$-grounded extension $Z$ of $F$ is empty. Otherwise, let $I_i$ be the set of initial arguments of $F$ $\mid _{B_{i-1}}$. We proceed to construct $Z$ by computing the sets $B_i$ as follows:
	

	\begin{compactenum}[1.~]
		\item $B_0 = \mathcal{A}$;
		\item $B_1 = B_0 \setminus (I_1 \cup R^+(I_1))$ and $Z=I_1$;
		\item \begin{compactenum}[(a)]
			\item compute $I_i \subseteq B_{i-1}$;
			\item $E_i = I_i \cap \mathbb{D}_w(Z)$, $Z = Z \cup E_i$, $F_i = I_i \setminus E_i$, $F_{i_0} = F_i$;
			\item $\forall a \in F_{i_j}$ (with $0 \leq j \leq |F_i|$), if $a \in \mathbb{D}_w(Z)$ then $Z = Z \cup \{a\}$ and $F_{i_{j+1}} = F_{i_j} \setminus \{a\}$;
			\item repeat (c) until $F_{i_j} = F_{i_{j-1}}$;
		\item $B_i = B_{i-1} \setminus \{I_i \cup R^+(I_i)\}$, with $2 \leq i \leq n$;
	\end{compactenum}
	\item repeat 3. until $B_i = \emptyset$ or $E_i = \emptyset$.
	\end{compactenum}

	This process can be done repeatedly until, for some $t$, $E_t = \emptyset$, where $2 \leq t \leq n$. From Theorem~\ref{th:union}, we know that the set union between $w$-admissible extensions is a $w$-admissible extension in turn. At this point, the set $Z = I_1 \cup E_2 \cup \dots \cup E_{t-1}$ is the $w$-grounded extension of $F$.	
	Note that the set $B_i$ coincides with the set of {\bfseries undec} arguments in the labelling of $B_{i-1}$ where $I_i$ is the set of {\bfseries in} arguments.

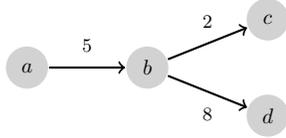
\begin{figure}[h]
		\centering
		\begin{tikzpicture}[scale=0.8, transform shape]
		\tikzstyle{every node} = [line width=1pt, shape=circle, fill=gray!35, minimum width=0.7cm]
		\node (a) at (0, -0.8) {$a$};
		\node (b) at (2, -0.8) {$b$};
		\node (c) at (4, 0) {$c$};
		\node (d) at (4, -1.6) {$d$};
		\draw [line width = 0.8pt, ->] (a) -- (b) node[pos=.5, fill=none, above] {\footnotesize{5}};
		\draw [line width = 0.8pt, ->] (b) -- (c) node[pos=.5, fill=none, above] {\footnotesize{2}};
		\draw [line width = 0.8pt, ->] (b) -- (d) node[pos=.5, fill=none, below] {\footnotesize{8}};
		\end{tikzpicture}
		\vspace{0.2cm}
		\caption{\label{fi:ex4}Example of a WAF with $\mathbb S = \mathbb S_\mathit{weighted}$.}
	\end{figure}

	\begin{example}\label{ex4}
		Let $F=\langle \mathcal{A}, R, W, \mathbb S \rangle$ be a WAF as in Figure~\ref{fi:ex4}. We have $I_1 = \{a\} \neq \emptyset$, so we look for the sets $B_i$. $B_1 = \mathcal{A} \setminus \{a,b\} = \{c,d\}$, so $I_2 = \{c,d\}$, $E_2 = \{c\}$ and $F_2 = \{d\}$. Consider $B_2 = \{c,d\} \setminus \{c,d\} = \emptyset$ that implies $E_3 = \emptyset$. $Z = \{a\} \cup \{c\} = \{a,c\}$ is the $w$-grounded extension of $F$.
	\end{example}

	\subsection*{Building w-preferred extensions}
	A $w$-preferred extension can be built incrementally by starting from some $w$-admissible extension.  Since the $w$-preferred semantics admits more extensions, different $w$-preferred extensions can be built, depending on both the initial extension and the selection of the nonempty $w$-admissible on each step of the procedure. Let $Z_i$ be any $w$-admissible extension of $F \mid _{B_{i-1}}$ and compute:
	
	\begin{compactenum}[1.~]
		\item $B_0 = \mathcal{A}$;
		\item $B_1 = B_0 \setminus (Z_1 \cup R^+(I_1))$ and $Z=Z_1$;
		\item \begin{compactenum}[(a)]
			\item compute $Z_i \subseteq B_{i-1}$;
			\item $E_i = Z_i \cap \mathbb{D}_w(Z)$, $Z = Z \cup E_i$, $F_i = Z_i \setminus E_i$, $F_{i_0} = F_i$;
			\item $\forall a \in F_{i_j}$ (with $0 \leq j \leq |F_i|$), if $a \in \mathbb{D}_w(Z)$ then $Z = Z \cup \{a\}$ and $F_{i_{j+1}} = F_{i_j} \setminus \{a\}$;
			\item repeat (c) until $F_{i_j} = F_{i_{j-1}}$;
			\item $B_i = B_{i-1} \setminus \{Z_i \cup R^+(Z_i)\}$, with $2 \leq i \leq n$;
		\end{compactenum}
		\item repeat 3. until $B_i = \emptyset$ or $E_i = \emptyset$.
	\end{compactenum}

	This process can be done repeatedly until, for some $t$, $E_t = \emptyset$, where $2 \leq t \leq n$. At this point, by Theorem~\ref{th:union}, the set $Z = Z_1 \cup E_2 \cup \dots \cup E_{t-1}$ is the $w$-preferred extension of $F$.
	
	\begin{example}
		Let $F=\langle \mathcal{A}, R, W, \mathbb S \rangle$ be a WAF as in Figure~\ref{fi:ex4}. Let's consider the $w$-admissible extension $Z_1 = \{a\}$ of $F$. Thus $B_1 = \{c,d\}$, $E_2 = \{c\}$ and $F_2 = \{d\}$. Since $B_2 = \emptyset$ and $E_3 = \emptyset$, $Z = \{a\} \cup \{c\} = \{a,c\}$ is the $w$-preferred extension of $F$.
	\end{example}

\noindent\textbf{Computational Complexity.} We analysed the above described algorithms from the computational point of view. The first algorithm, which computes $w$-grounded extensions, has an overall time complexity of $O(n^4)$. The algorithm for $w$-preferred extensions reveals worse performance than the first one, with a time complexity of $O(2^n \cdot n^5)$. This is due to the fact that an admissible extension has to be found at each execution of step 3. A more extended study of the complexity is left for future work.

\section{Conclusion and Future Work}\label{se:conclusions}
	In this work, we introduce a matrix approach for studying extensions of semiring-based semantics. A WAF is represented as a matrix in which all elements correspond to weights assigned to relations among arguments. In particular, by extracting sub-blocks from this matrix, it is possible to check if a set of arguments is an extension for some semantics.
	Also, we describe an incremental procedure for building $w$-grounded and $w$-preferred extensions and we study how to reduce the number of arguments of a WAF in order to obtain a contracted matrix with the same status as the original one (w.r.t.\ the semantics). A possible application for this approach could be the identification of equational representation of semiring-based extensions, by using the method proposed in \cite{DBLP:conf/ecsqaru/Gabbay11}. We plan to extend our current implementation\footnote{http://www.dmi.unipg.it/conarg}~\cite{DBLP:conf/ictai/BistarelliS11,DBLP:conf/tafa/BistarelliS11} with the proposed approaches, and to test their performance on real applications. Finally, we would like to investigate whether such methodologies can be applied when considering coalitions of arguments~\cite{DBLP:journals/fuin/BistarelliS13}.

\bibliographystyle{aaai}
\bibliography{biblio}

\begin{thebibliography}{}

\bibitem[\protect\citeauthoryear{Bistarelli and
  Santini}{2011a}]{DBLP:conf/ictai/BistarelliS11}
Bistarelli, S., and Santini, F.
\newblock 2011a.
\newblock {ConArg}: {A} constraint-based computational framework for
  argumentation systems.
\newblock In {\em {ICTAI} 2011},  605--612.
\newblock {IEEE} Computer Society.

\bibitem[\protect\citeauthoryear{Bistarelli and
  Santini}{2011b}]{DBLP:conf/tafa/BistarelliS11}
Bistarelli, S., and Santini, F.
\newblock 2011b.
\newblock Modeling and solving {AFs} with a constraint-based tool: {ConArg}.
\newblock In {\em {TAFA} 2011}, volume 7132 of {\em LNCS},  99--116.
\newblock Springer.

\bibitem[\protect\citeauthoryear{Bistarelli and
  Santini}{2013}]{DBLP:journals/fuin/BistarelliS13}
Bistarelli, S., and Santini, F.
\newblock 2013.
\newblock Coalitions of arguments: An approach with constraint programming.
\newblock {\em Fundam. Inform.} 124(4):383--401.

\bibitem[\protect\citeauthoryear{Bistarelli, Pirolandi, and
  Santini}{2010}]{DBLP:conf/cilc/BistarelliPS10}
Bistarelli, S.; Pirolandi, D.; and Santini, F.
\newblock 2010.
\newblock Solving weighted argumentation frameworks with soft constraints.
\newblock In {\em Proceedings of the 25th Italian Conference on Computational
  Logic, 2010}, volume 598 of {\em {CEUR} Workshop Proceedings}.
\newblock CEUR-WS.org.

\bibitem[\protect\citeauthoryear{Bistarelli, Rossi, and
  Santini}{2016}]{DBLP:conf/flairs/BistarelliRS16}
Bistarelli, S.; Rossi, F.; and Santini, F.
\newblock 2016.
\newblock A collective defence against grouped attacks for weighted abstract
  argumentation frameworks.
\newblock In {\em Proceedings of {FLAIRS} 2016},  638--643.
\newblock {AAAI} Press.

\bibitem[\protect\citeauthoryear{Dung}{1995}]{dung}
Dung, P.~M.
\newblock 1995.
\newblock On the acceptability of arguments and its fundamental role in
  nonmonotonic reasoning, logic programming and n-person games.
\newblock {\em Artif. Intell.} 77(2):321--357.

\bibitem[\protect\citeauthoryear{Dunne \bgroup et al\mbox.\egroup
  }{2011}]{DBLP:journals/ai/DunneHMPW11}
Dunne, P.~E.; Hunter, A.; McBurney, P.; Parsons, S.; and Wooldridge, M.
\newblock 2011.
\newblock Weighted argument systems: Basic definitions, algorithms, and
  complexity results.
\newblock {\em Artif. Intell.} 175(2):457--486.

\bibitem[\protect\citeauthoryear{Gabbay}{2011}]{DBLP:conf/ecsqaru/Gabbay11}
Gabbay, D.~M.
\newblock 2011.
\newblock Introducing equational semantics for argumentation networks.
\newblock In {\em Proceedings of {ECSQARU} 2011}, volume 6717 of {\em LNCS},
  19--35.
\newblock Springer.

\bibitem[\protect\citeauthoryear{Xu and Cayrol}{2015}]{xu15}
Xu, Y., and Cayrol, C.
\newblock 2015.
\newblock The matrix approach for abstract argumentation frameworks.
\newblock In {\em Proceedings of {TAFA} 2015}, volume 9524 of {\em LNCS},
  243--259.
\newblock Springer.

\end{thebibliography}

\end{document}